\documentclass[lettersize,journal]{IEEEtran}
\usepackage{amsmath,amsfonts}
\usepackage{algorithmic}
\usepackage{algorithm}
\usepackage{array}
\usepackage{textcomp}
\usepackage{booktabs}
\usepackage{subcaption}
\usepackage{stfloats}
\usepackage{url}
\usepackage{verbatim}
\usepackage{graphicx}
\usepackage{cite}

\usepackage{comment}
\usepackage{amsthm}
\usepackage{pifont}  % for xmarks
\usepackage{graphicx} % for emojis if needed
\usepackage{url}
\usepackage{booktabs} % For formal tables
\usepackage{xspace}
\usepackage{graphicx}
\usepackage{makecell}
\usepackage{amssymb}
\usepackage{diagbox}
\usepackage{algorithm}
\usepackage{algorithmic}
\usepackage{listings}
\usepackage{amsmath}
\usepackage{graphics}
\usepackage{epsfig}
\usepackage{color}
\usepackage{ifpdf}
\usepackage{mathrsfs}
\usepackage{algorithm}
\usepackage{algorithmic}
\usepackage{multirow}
\usepackage{verbatim}
\usepackage{xcolor}
\usepackage{url}
\usepackage{color,xcolor}
\usepackage{array}
\usepackage{multirow}
\usepackage{bm}
\usepackage{CJKutf8}
\usepackage{longtable}
\usepackage{tcolorbox}
\usepackage[normalem]{ulem}

\begin{document}

\title{Design and Evaluation of Deep Learning-Based Dual-Spectrum Image Fusion Methods}

\author{BeiNing Xu, ~\IEEEmembership{School of Automation, Shanghai Jiao Tong University}, \\
Junxian Li,~\IEEEmembership{School of Computer Science, Shanghai Jiao Tong University}}
        % <-this % stops a space
%\thanks{This paper was produced by the IEEE Publication Technology Group. They are in Piscataway, NJ.}% <-this % stops a space
%\thanks{Manuscript received April 19, 2021; revised August 16, 2021.}}

% The paper headers
\markboth{Journal of \LaTeX\ Class Files,~Vol.~14, No.~8, August~2021}%
{Shell \MakeLowercase{\textit{et al.}}: A Sample Article Using IEEEtran.cls for IEEE Journals}

%\IEEEpubid{0000--0000/00\$00.00~\copyright~2021 IEEE}
% Remember, if you use this you must call \IEEEpubidadjcol in the second
% column for its text to clear the IEEEpubid mark.

\maketitle

\begin{abstract}
    Visible images offer rich texture details, while infrared images emphasize salient targets. Fusing these complementary modalities enhances scene understanding, particularly for advanced vision tasks under challenging conditions. Recently, deep learning-based fusion methods have gained attention, but current evaluations primarily rely on general-purpose metrics without standardized benchmarks or downstream task performance. Additionally, the lack of well-developed dual-spectrum datasets and fair algorithm comparisons hinders progress.

To address these gaps, we construct a high-quality dual-spectrum dataset captured in campus environments, comprising 1,369 well-aligned visible–infrared image pairs across four representative scenarios: daytime, nighttime, smoke occlusion, and underpasses. We also propose a comprehensive and fair evaluation framework that integrates fusion speed, general metrics, and object detection performance using the lang-segment-anything model to ensure fairness in downstream evaluation.

Extensive experiments benchmark several state-of-the-art fusion algorithms under this framework. Results demonstrate that fusion models optimized for downstream tasks achieve superior performance in target detection, especially in low-light and occluded scenes. Notably, some algorithms that perform well on general metrics do not translate to strong downstream performance, highlighting limitations of current evaluation practices and validating the necessity of our proposed framework.

The main contributions of this work are: (1) a campus-oriented dual-spectrum dataset with diverse and challenging scenes; (2) a task-aware, comprehensive evaluation framework; and (3) thorough comparative analysis of leading fusion methods across multiple datasets, offering insights for future development.
\end{abstract}


\section{Introduction}

\begin{figure}[htbp]
    \centering
    \includegraphics[width=1\linewidth]{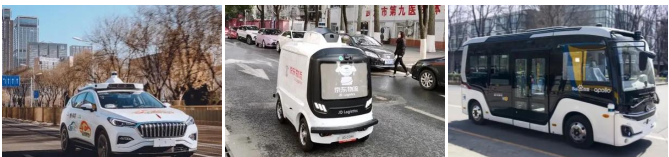}
    \caption{Usage of autonomous driving.}
    \label{fig:usage_1}
\end{figure}

\begin{figure}[htbp]
    \centering
    \includegraphics[width=1\linewidth]{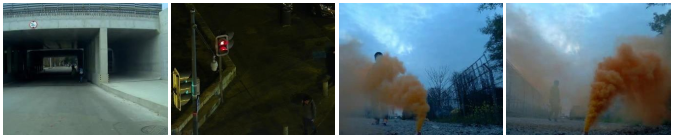}
    \caption{Examples of special cases.}
    \label{fig:special_cases}
\end{figure}

In recent years, the rapid development of autonomous driving technologies has led to the deployment of self-driving taxis, unmanned delivery vehicles, and autonomous buses~\ref{fig:usage_1}. 
A core component of such systems is the perception module, which enables environmental understanding and decision-making. However, perception performance is often challenged in adverse conditions such as low illumination, strong glare, and occlusion, especially in real-world scenarios like nighttime roads, smoke or fog, and transitional lighting environments (e.g., tunnels or underpasses), as shown in Fig.~\ref{fig:special_cases}.

\begin{figure}
    \centering
    \includegraphics[width=0.75\linewidth]{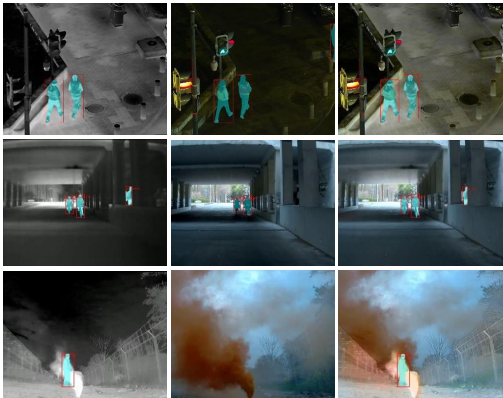}
    \caption{Examples of fused images.}
    \label{fig:fusion_cases}
\end{figure}

To address the limitations of single-modal sensors, researchers have explored multi-modal sensing strategies, especially the fusion of visible and infrared (IR) images. Visible images offer rich texture and structural details but degrade under poor lighting. In contrast, infrared images highlight thermal targets and maintain robustness in low-light and occluded conditions but lack fine textures and colors. As illustrated in Fig.~\ref{fig:fusion_cases}, 
combining these complementary modalities into a single fused image can provide a more comprehensive and discriminative representation of the scene, benefiting both human interpretation and machine perception.

Image fusion, especially between visible and infrared images, has therefore become a key enabler of robust perception in autonomous systems. Recent advances in deep learning have significantly improved fusion quality through sophisticated feature extraction and attention mechanisms, including CNN-, Transformer-, and GAN-based architectures. Despite these advances, the current literature faces three major challenges:

\begin{itemize}
\item \textbf{Insufficient and inconsistent evaluation protocols:} Fusion performance is often assessed using diverse general-purpose metrics such as entropy, mutual information, or structural similarity. However, these metrics are not standardized across works and fail to reflect performance in downstream tasks like object detection.
\item \textbf{Lack of high-quality, scenario-specific datasets:} Most existing dual-spectral datasets target urban driving scenes and lack coverage of more complex environments such as campus roads, which feature long-range small objects, high pedestrian density, bicycles or e-bikes, and frequent occlusions. In addition, challenging conditions like smoke and tunnel-like lighting transitions are rarely captured.
\item \textbf{Neglect of downstream task performance:} The core motivation of fusion is to enhance perception for real-world applications. Yet many works focus solely on pixel-level fusion quality, ignoring how well the fused images support downstream tasks such as pedestrian detection.
\end{itemize}

To address these limitations, this paper proposes a comprehensive framework encompassing dataset creation, evaluation system design, and algorithm benchmarking:

\begin{itemize}
\item We construct a dual-spectral dataset collected in real campus environments. The dataset contains 1,369 well-aligned visible-infrared image pairs covering four representative scenarios: daytime, nighttime, smoke occlusion, and transitional lighting (e.g., under bridges). The dataset also includes detection annotations in selected scenes to support downstream evaluation.
\item We design a unified and fair evaluation system that includes three components: fusion speed, general fusion quality metrics (e.g., entropy, PSNR, SSIM), and downstream performance using a powerful zero-shot detection pipeline—\texttt{lang-segment-anything}—built upon Grounding DINO and Segment Anything Model\cite{liu2024grounding,kirillov2023segment}.
\item We conduct extensive experiments across multiple datasets, comparing several state-of-the-art fusion methods under both general metrics and detection accuracy. Our results highlight the limitations of traditional metrics and demonstrate the importance of downstream-aware fusion evaluation.
\end{itemize}

In summary, our work contributes both a high-quality, scenario-diverse dataset and a comprehensive benchmarking pipeline that bridges the gap between low-level fusion and high-level perception. This effort aims to guide future research toward more practical and task-effective fusion algorithms.

\section{Related Works}

Early research on infrared and visible image fusion mainly focused on traditional methods in the spatial or transform domains. These methods manually designed fusion rules based on activity-level measurements, often neglecting the modality-specific characteristics of the input images. As a result, their performance was limited in complex real-world scenarios~\cite{kaur2021image}.

In recent years, deep learning-based fusion algorithms have significantly advanced the field. Convolutional neural network (CNN)-based methods extract features from different modalities using parallel or shared branches and learn fusion strategies through end-to-end optimization. Several works have enhanced fusion quality by designing cross-modal attention mechanisms~\cite{liu2025pafusion}, semantic-aware loss functions~\cite{tang2022image}, and lightweight architectures suitable for real-time applications~\cite{sun2022detfusion}.

Autoencoder (AE)-based approaches first train modality-specific encoders and decoders to extract deep features and then apply hand-crafted or learned strategies for feature fusion. Although these methods retain informative latent representations, they are often not fully end-to-end and may struggle with generalization~\cite{jian2020sedrfuse, li2022mafusion}.

Generative adversarial networks (GANs) have also been explored for unsupervised fusion by aligning the distribution of fused outputs with that of the source images. While GAN-based methods~\cite{liu2022target} can enhance perceptual quality and semantic alignment, training instability and mode collapse remain challenges.

Recently, Transformer-based methods have drawn increasing attention due to their capability to model global dependencies and cross-modal correlations. Hybrid CNN-Transformer architectures such as SwinFusion~\cite{ma2022swinfusion} and CDDFuse~\cite{zhao2023cddfuse} combine local and global feature representations and have achieved strong performance. More advanced variants like YDTR~\cite{tang2022ydtr} and PSFusion~\cite{tang2023rethinking} integrate progressive semantic injection and task-specific constraints to further improve fusion for downstream applications.

Despite these advances, most existing works focus only on general fusion metrics and do not consider real-world performance in high-level vision tasks, motivating our work to bridge this gap through task-aware evaluation.
\section{Methodology}

This section describes the construction of the dual-spectral dataset in campus environments and the design of a unified evaluation system for infrared-visible image fusion. The proposed framework focuses on practical deployment scenarios such as autonomous navigation in adverse lighting or occluded conditions and is tailored to both general-purpose fusion quality and downstream vision task performance.

\subsection{Campus Dual-Spectral Dataset Construction}
\label{sec:dataset}

\subsubsection{Dataset Overview}

\begin{figure}
    \centering
    \includegraphics[width=\linewidth]{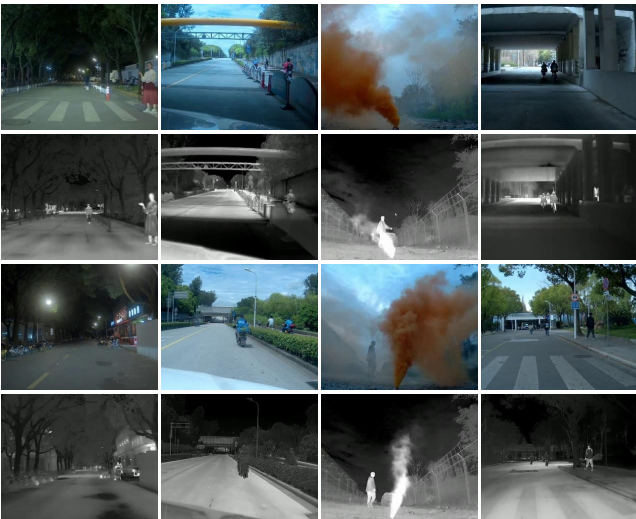}
    \caption{Dataset Scenarios.}
    \label{fig:dataset_scenarios}
\end{figure}

Existing dual-spectral datasets are often focused on urban driving scenes and lack diversity in campus settings, where long-range small objects, smoke, and lighting transitions (e.g., underpasses) frequently occur. To fill this gap, we construct a well-aligned and high-resolution dual-spectral dataset, as illustrated in Fig.~\ref{fig:dataset_scenarios}.
The dataset comprises 1,369 pairs of visible and infrared images with a resolution of 640×512, covering four typical scenarios: daytime, nighttime, smoke occlusion, and underpass illumination changes. Table~\ref{tab:dataset_summary} summarizes the dataset statistics.

\begin{table}[!ht]
\caption{Summary of Campus Dual-Spectral Dataset}
\label{tab:dataset_summary}
\centering
\begin{tabular}{lccc}
\toprule
\textbf{Scenario} & \textbf{Image Pairs} & \textbf{Aligned} & \textbf{Annotated} \\
\midrule
Daytime Campus Road & 159 & Yes & Yes \\
Nighttime Campus Road & 80 & Yes & Yes \\
Smoke Occlusion & 550 & Yes & No \\
Underpass Lighting & 580 & Yes & No \\
\bottomrule
\end{tabular}
\end{table}

\subsubsection{Data Acquisition and Pre-processing}

The raw data were recorded as temporally synchronized visible and infrared videos at 10 FPS using dual-modality cameras. However, due to varying field-of-views (FoV), geometric misalignment exists. We perform alignment in two stages:

Initial Calibration: Intrinsic and extrinsic camera parameters are estimated to compute a homography matrix for coarse registration.

Deep Alignment: We use SuperFusion~\cite{tang2022superfusion} to obtain pixel-level alignment. SuperFusion models registration and fusion jointly by learning a bidirectional deformation field optimized under photometric and endpoint constraints.

\begin{figure}[htbp]
    \centering
    \begin{subfigure}{1\linewidth}
        \includegraphics[width=1\linewidth]{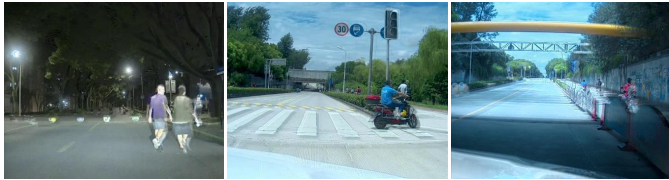}
        \subcaption{Fusion results before alignment.}
        \label{fig:before_align}
    \end{subfigure}
    \begin{subfigure}{1\linewidth}
        \centering
        \includegraphics[width=1\linewidth]{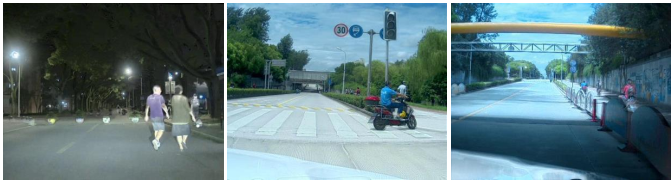}
        \subcaption{Fusion results after alignment.}
        \label{fig:after_align}
    \end{subfigure}
    \caption{Comparison between before and after alignment.}
    \label{fig:alignment_comparison}
\end{figure}

Fig.~\ref{fig:alignment_comparison} compares fusion results before and after alignment. To support object detection tasks, images from the daytime and nighttime road scenes are annotated using LabelImg in YOLO format, with bounding boxes marking pedestrian targets.

\subsection{Fusion Evaluation System Design}

To comprehensively evaluate image fusion performance, we propose a three-pronged evaluation protocol that measures: (1) fusion speed, (2) general-purpose image fusion metrics, and (3) performance on downstream tasks.

\subsubsection{Fusion Speed}

We report the average runtime (in seconds) of the fusion operation per image pair, excluding model loading and I/O latency. This metric reflects the algorithm’s real-time feasibility in embedded systems.

\subsubsection{General Fusion Metrics}

We adopt the following widely-used metrics to quantify fusion quality:

\begin{itemize}
\item \textbf{Entropy (EN)}:
\[
EN = -\sum_{i=1}^{L} p_i \log_2(p_i)
\]
where $p_i$ denotes the probability of grayscale level $i$, and 
$L$ is the number of gray levels.

\item \textbf{Standard Deviation (SD)}:
\[
SD = \sqrt{\frac{1}{MN} \sum_{x=1}^{M} \sum_{y=1}^{N} (F(x,y) - \bar{F})^2}
\]
where \(\bar{F}\) is the mean intensity of fused image \(F\).

\item \textbf{Mutual Information (MI)}:
\[
MI = \sum_{a=1}^{L} \sum_{b=1}^{L} P_{AB}(a,b) \log \left( \frac{P_{AB}(a,b)}{P_A(a)P_B(b)} \right)
\]
where \(P_{AB}(a,b)\) is the joint histogram of the source and fused images, and \(P_A\), \(P_B\) are marginal histograms.

\item \textbf{Peak Signal-to-Noise Ratio (PSNR)}:
\[
PSNR = 10 \log_{10} \left( \frac{MAX^2}{MSE} \right)
\]
where \(MSE = \frac{1}{MN} \sum_{i,j} (F(i,j) - I(i,j))^2\), and \(MAX\) is the maximum possible pixel value (e.g., 255).

\item \textbf{Edge Preservation (Q\_abf)}~\cite{charbonnier1997deterministic}:
\[
Q_{abf} = \frac{w_A Q_A + w_B Q_B}{w_A + w_B}
\]
where \(Q_A\), \(Q_B\) measure edge similarity between the fused and source images, and \(w_A\), \(w_B\) are edge strength weights.

\item \textbf{Structural Similarity Index (SSIM)}:
\[
SSIM(A,F) = \frac{(2\mu_A\mu_F + C_1)(2\sigma_{AF} + C_2)}{(\mu_A^2 + \mu_F^2 + C_1)(\sigma_A^2 + \sigma_F^2 + C_2)}
\]
where \(\mu\) is mean, \(\sigma\) is standard deviation, \(\sigma_{AF}\) is covariance between \(A\) and \(F\), and \(C_1\), \(C_2\) are constants.
\end{itemize}

\subsubsection{Downstream Task Evaluation: Object Detection}

To evaluate the utility of fused images in high-level vision tasks, we test object detection performance using the Lang-Segment-Anything (LSA) framework. LSA integrates Grounding-DINO~\cite{liu2024grounding} for open-set text-driven object detection and SAM~\cite{kirillov2023segment} for mask generation. This pipeline enables zero-shot pedestrian detection using text prompts such as "a person".

\begin{figure}
    \centering
    \includegraphics[width=1\linewidth]{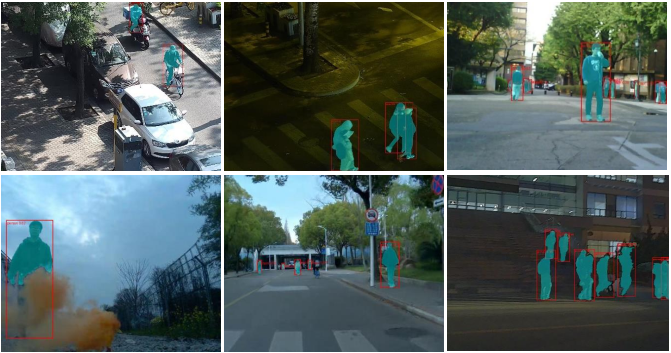}
    \caption{Detection examples on fused images.}
    \label{fig:lsa_results}
\end{figure}

Compared to traditional detection models (e.g., YOLO), which require modality-specific training, LSA offers fair comparison across modalities, as it does not require model retraining for each input type. Fig.~\ref{fig:lsa_results} demonstrates detection examples on fused versus single-modality images.

We report standard detection accuracy using \textbf{mean Average Precision at 0.5 IoU (mAP@0.5)} as the primary quantitative metric.

\section{Experiments}

\subsection{Experiment Settings}

\textbf{Datasets.} We evaluate our method on three datasets, including a newly collected campus-based dual-spectrum dataset, the public MSRS dataset, and the LLVIP dataset. Each dataset offers distinct scene characteristics and complements the evaluation across different environments and task settings.

The LLVIP dataset~\cite{jia2021llvip} focuses on low-light urban street scenes, featuring a large number of annotated pedestrians. It is widely used for vision tasks under low illumination, such as image fusion, pedestrian detection, and image translation. The dataset provides aligned visible and infrared image pairs along with pixel-level annotations, making it especially suitable for fusion methods targeting real-world safety-critical applications.

The MSRS dataset~\cite{tang2022image} covers both daytime and nighttime urban scenes with well-aligned visible and infrared image pairs. The infrared images are preprocessed for improved clarity; however, the dataset lacks annotations for high-level vision tasks. Despite this, it has been frequently adopted in recent image fusion literature due to its diverse lighting conditions and alignment quality.

In addition, we contribute a campus dual-spectrum dataset, specifically captured to reflect diverse campus environments across varying lighting and seasonal conditions. The dataset includes well-aligned high-resolution image pairs and is intended to support comprehensive evaluation in academic settings. Details of this dataset are provided in Section~\ref{sec:dataset}.

\textbf{Baselines.} We selected six recent and representative fusion methods that have shown superior performance in visible-infrared image fusion. The selected baselines include DetFusion~\cite{sun2022detfusion}, SeAFusion~\cite{tang2022image}, SwinFusion~\cite{ma2022swinfusion}, CDDFuse~\cite{zhao2023cddfuse}, PSFusion~\cite{tang2023rethinking}, and PIAFusion~\cite{tang2022piafusion}.

\textbf{Hyper-parameters.} We conduct all our experiments on a single RTX 4090, with memory of 24GB. We follow the initial hyper-parameters described in their paper, and train them each for 50-100 epochs until convergence.

\textbf{Evaluation Metrics.} All our evaluation metrics are listed in our methodology section. 

\subsection{Baseline Details}

Based on a comprehensive literature review, we observe that recent visible-infrared image fusion approaches predominantly adopt CNN-based or Transformer-based architectures, whereas relatively few methods are based on GANs or Autoencoders. To ensure a representative and competitive comparison, we select six state-of-the-art fusion methods as baselines, including four CNN-based and two Transformer-based approaches.

Among the CNN-based methods, PSFusion, PIAFusion, DetFusion, and SeAFusion have demonstrated strong fusion performance. Notably, DetFusion and SeAFusion incorporate high-level vision tasks into the fusion pipeline to enhance task-specific performance.

DetFusion~\cite{sun2022detfusion} integrates object detection with image fusion through a shared SA-Fusion branch and dual detection branches, where detection losses guide the fusion network's training.

SeAFusion~\cite{tang2022image} introduces a lightweight gradient residual dense block (GRDB) network for feature fusion, and leverages semantic segmentation losses to supervise the fusion process.

PIAFusion~\cite{tang2022piafusion} proposes a cross-modality differential-aware fusion (CMDAF) module within a CNN-based backbone, using illumination-aware losses to enhance complementary information integration.

PSFusion~\cite{tang2023rethinking} adopts a dual-branch architecture, consisting of a scene reconstruction path and a sparse semantic-aware path. It utilizes a progressive semantic injection module (PSIM) and a deep semantic fusion module (PSFM) to embed semantic information and guide feature integration.

The Transformer-based methods include SwinFusion and CDDFuse, both of which effectively capture global dependencies:

SwinFusion~\cite{ma2022swinfusion} combines CNN and Swin Transformer~\cite{liu2021swin} to jointly capture local and global features and performs cross-modality fusion using Transformer layers.

CDDFuse~\cite{zhao2023cddfuse} employs Restormer~\cite{zamir2022restormer} blocks for shallow feature extraction, and utilizes INN and Lite Transformer modules to separately extract high- and low-frequency features, followed by local (INN) and global (LT) fusion strategies.

These baselines reflect the current trends and design philosophies in image fusion research and provide a strong foundation for comparative evaluation.

\subsection{Qualitative Results}

The following figures present a qualitative comparison of fusion results.

\subsubsection{Results on MSRS dataset}
\begin{figure*}
    \centering
    \includegraphics[width=1\linewidth]{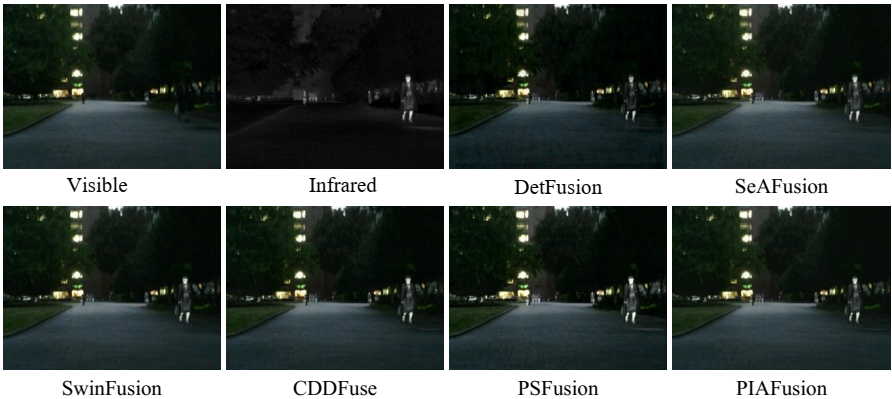}
    \caption{Results on MSRS dataset (night).}
    \label{fig:msrs_night}
\end{figure*}

As shown in Fig.~\ref{fig:msrs_night}, on the MSRS dataset, all selected methods exhibit a certain degree of complementary feature preservation from visible and infrared modalities. Specifically, PSFusion effectively retains the salient contrast from infrared images, highlighting targets while preserving fine-grained textures from the visible images. In the region near the tree branches in front of the house, it can be observed that while PIAFusion and SwinFusion emphasize the salient targets, they lose some structural and texture details.

\begin{figure*}
    \centering
    \includegraphics[width=1\linewidth]{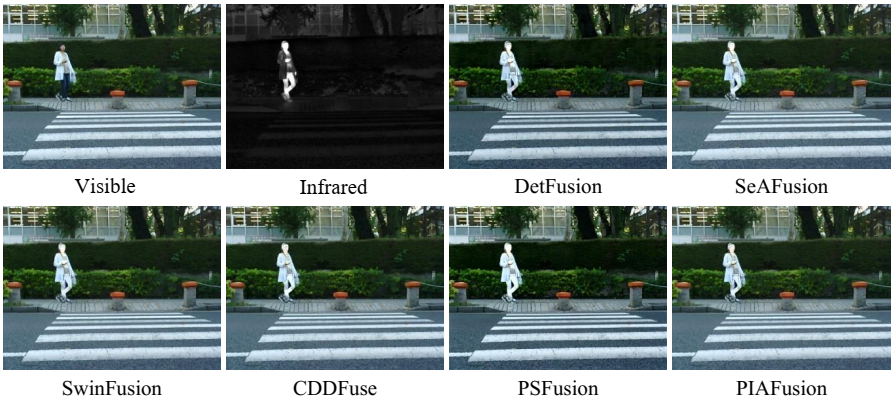}
    \caption{Results on MSRS dataset (day).}
    \label{fig:msrs_day}
\end{figure*}

In Fig.~\ref{fig:msrs_day}, which shows fusion results under daytime conditions, all methods successfully preserve the salient intensity contrast of the infrared modality. Notably, CDDFuse and PIAFusion demonstrate superior ability in maintaining texture details of salient targets—such as hair and clothing of pedestrians—originating from the visible images.

\subsubsection{Results on Strong Glare Cases}
\begin{figure*}
    \begin{subfigure}{1\linewidth}
    \centering
    \includegraphics[width=1\linewidth]{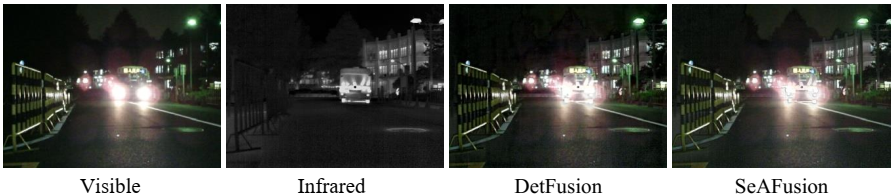}
    \label{fig:glare_1}
    \end{subfigure}
    \begin{subfigure}{1\linewidth}
        \centering
        \includegraphics[width=1\linewidth]{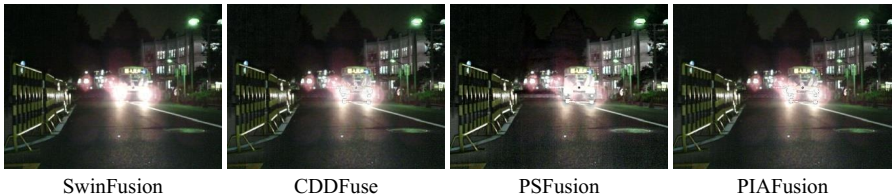}
        \label{fig:glare_2}
    \end{subfigure}
    \caption{Results under glare situation.}
    \label{fig:glare_case}
\end{figure*}

In strong glare scenarios caused by direct lighting into the camera lens, targets located behind the light source are often difficult to distinguish in visible images. As shown in Fig.~\ref{fig:glare_case}, under direct illumination from vehicle headlights, DetFusion and SwinFusion perform similarly to the visible image and fail to highlight the outline of the bus effectively. In contrast, the other methods are able to better preserve and enhance the contours of the bus headlights and tires, demonstrating superior robustness under glare conditions.

\subsubsection{Results on All-weather Campus Dataset}
\begin{figure*}
    \centering
    \includegraphics[width=1\linewidth]{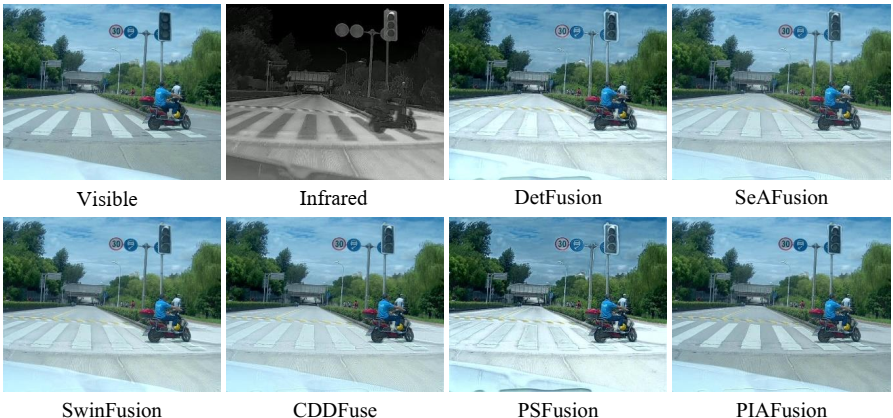}
    \caption{Results on Campus (day).}
    \label{fig:campus_day}
\end{figure*}

\begin{figure*}
    \centering
    \includegraphics[width=1\linewidth]{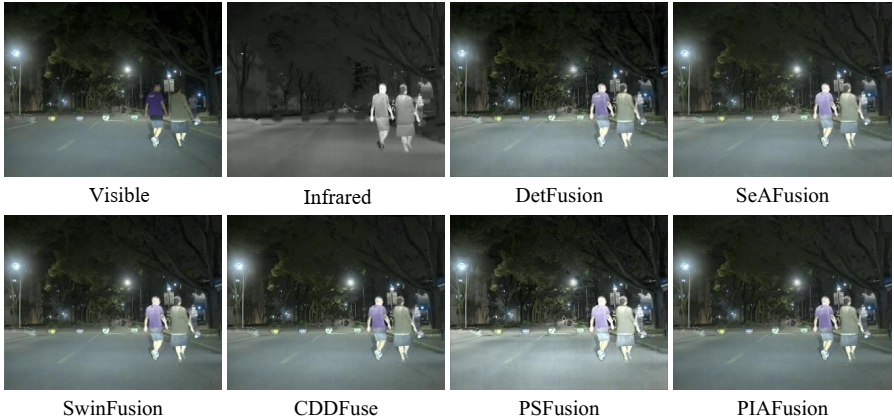}
    \caption{Results on Campus (night).}
    \label{fig:campus_night}
\end{figure*}

On the all-weather campus dual-spectrum dataset, as illustrated in Fig.~\ref{fig:campus_day} and Fig.~\ref{fig:campus_night}, all baseline methods demonstrate a certain ability to integrate thermal radiation information from infrared images with fine texture details from visible images. PSFusion consistently maintains high contrast but suffers from less distinct edges and exhibits a degree of overexposure under daytime conditions.
CDDFuse and SeAFusion show higher similarity to the source images while better preserving edge structures.
Additionally, in the nighttime sky regions, both DetFusion and PSFusion introduce noticeable noise originating from the infrared modality.

\subsubsection{Results on LLVIP dataset.}
\begin{figure*}
    \centering
    \includegraphics[width=0.75\linewidth]{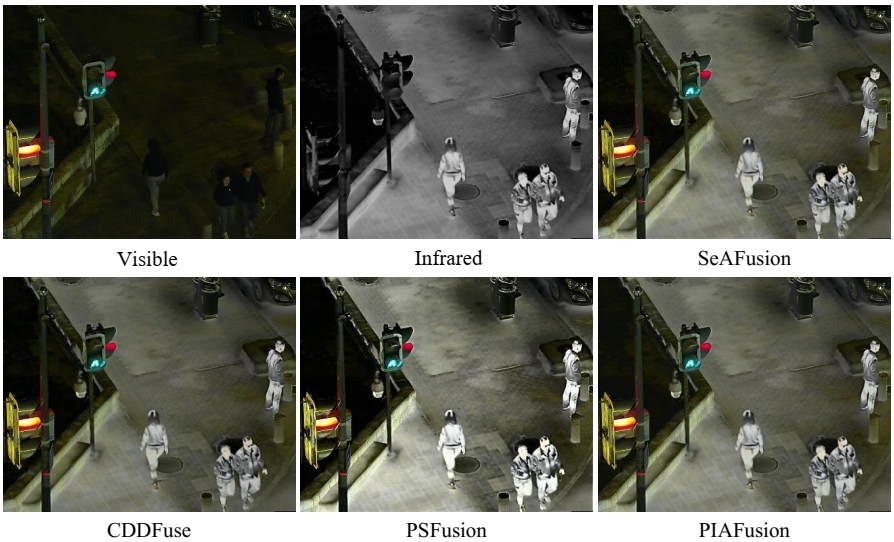}
    \caption{Results on the LLVIP dataset.}
    \label{fig:llvip_results}
\end{figure*}

On the LLVIP dataset, as shown in Fig.~\ref{fig:llvip_results}, the fused images effectively highlight pedestrian targets under extremely low-light conditions while preserving the color information from visible images.
By examining the ground textures, it is evident that PSFusion preserves texture details more thoroughly and exhibits the most prominent enhancement of salient targets.
Meanwhile, CDDFuse demonstrates strong capability in integrating edge and contour information from both source modalities.

\subsection{Quantitative Results}

We quantitatively evaluate the performance of fusion algorithms using multiple metrics, including fusion speed, information entropy (EN), standard deviation (SD), mutual information (MI), peak signal-to-noise ratio (PSNR), edge-based evaluation metrics, and structural similarity index (SSIM).

\subsubsection{Results on Campus Dataset}
On the proposed campus dual-spectrum dataset, we test all baseline methods in terms of fusion speed and the above standard evaluation metrics. The experimental results are summarized in Table~\ref{tab:quantitative_campus}.
In the table, red indicates the best performance and green denotes the second best. Each value is reported in the format of mean ± standard deviation.
\begin{table*}[htbp]
\caption{Performance Comparison of Fusion Methods on the Campus Dual-Spectrum Dataset}
\label{tab:quantitative_campus}
\centering
\begin{tabular}{lcccccc}
\toprule
\textbf{Method} & \textbf{DetFusion} & \textbf{SeAFusion} & \textbf{SwinFusion} & \textbf{CDDFuse} & \textbf{PSFusion} & \textbf{PIAFusion} \\
\midrule
Speed $\downarrow$ & 0.108 & \textcolor{red}{0.032} & 2.127 & 0.429 & 0.075 & \textcolor{green}{0.050} \\
EN $\uparrow$ & 7.364$\pm$0.258 & \textcolor{green}{7.472$\pm$0.246} & 7.348$\pm$0.259 & 7.362$\pm$0.248 & \textcolor{red}{7.578$\pm$0.134} & 7.335$\pm$0.267 \\
SD $\uparrow$ & 58.113$\pm$10.78 & \textcolor{green}{65.57$\pm$12.382} & 58.319$\pm$10.61 & 61.529$\pm$11.228 & \textcolor{red}{73.504$\pm$11.609} & 54.496$\pm$9.463 \\
MI $\uparrow$ & \textcolor{red}{4.533$\pm$0.43} & 3.488$\pm$0.362 & 4.028$\pm$0.384 & 4.170$\pm$0.347 & 3.537$\pm$0.352 & \textcolor{green}{4.175$\pm$0.374} \\
PSNR $\uparrow$ & \textcolor{red}{62.987$\pm$3.651} & \textcolor{green}{62.713$\pm$3.144} & 62.615$\pm$3.442 & 62.544$\pm$3.162 & 60.131$\pm$3.482 & 63.148$\pm$3.343 \\
Qabf $\uparrow$ & \textcolor{green}{0.492$\pm$0.030} & \textcolor{red}{0.493$\pm$0.025} & 0.463$\pm$0.030 & 0.490$\pm$0.028 & 0.473$\pm$0.025 & 0.482$\pm$0.035 \\
SSIM $\uparrow$ & \textcolor{red}{0.952$\pm$0.037} & \textcolor{green}{0.832$\pm$0.039} & 0.878$\pm$0.037 & 0.934$\pm$0.033 & 0.800$\pm$0.040 & 0.860$\pm$0.040 \\
\bottomrule
\end{tabular}
\end{table*}
From the results, it can be observed that SeAFusion achieves the fastest fusion speed (0.032 s), followed closely by PIAFusion (0.050 s). In terms of information richness and visual quality, PSFusion achieves the highest EN (7.578) and SD (73.504), indicating stronger contrast and richer details, but at the cost of some overexposure observed in qualitative results. DetFusion leads in mutual information (MI) and PSNR, suggesting better preservation of complementary information and higher reconstruction fidelity. For perceptual quality metrics, PSFusion and SeAFusion achieve the highest Qabf scores, while DetFusion obtains the best SSIM score (0.952), reflecting its superior structural preservation.

\subsubsection{Results on Open-source Datasets}
\begin{table*}[htbp]
\caption{Performance Comparison of Fusion Methods on the MSRS Dataset}
\label{tab:msrs_results}
\centering
\begin{tabular}{lcccccc}
\toprule
\textbf{Method} & \textbf{DetFusion} & \textbf{SeAFusion} & \textbf{SwinFusion} & \textbf{CDDFuse} & \textbf{PSFusion} & \textbf{PIAFusion} \\
\midrule
Speed $\downarrow$ & 0.234 & \textcolor{red}{0.201} & 2.680 & 0.458 & 0.264 & \textcolor{green}{0.223} \\
EN $\uparrow$ & 6.492$\pm$0.953 & 6.651$\pm$0.731 & 6.621$\pm$0.780 & 6.699$\pm$0.745 & \textcolor{red}{6.838$\pm$0.696} & \textcolor{green}{6.571$\pm$0.842} \\
SD $\uparrow$ & 41.653$\pm$13.622 & 41.843$\pm$12.754 & 42.982$\pm$13.241 & \textcolor{green}{43.368$\pm$13.685} & \textcolor{red}{46.279$\pm$12.595} & 42.536$\pm$13.311 \\
MI $\uparrow$ & 2.563$\pm$0.916 & 4.031$\pm$0.898 & 4.531$\pm$1.250 & \textcolor{red}{4.951$\pm$1.125} & 2.948$\pm$0.450 & \textcolor{green}{4.575$\pm$1.171} \\
PSNR $\uparrow$ & \textcolor{red}{65.26$\pm$4.569} & 64.331$\pm$4.545 & \textcolor{green}{64.451$\pm$4.725} & 64.412$\pm$4.948 & 64.067$\pm$3.899 & 64.305$\pm$4.553 \\
Qabf $\uparrow$ & 0.638$\pm$0.074 & 0.675$\pm$0.058 & 0.654$\pm$0.077 & \textcolor{red}{0.689$\pm$0.065} & 0.675$\pm$0.049 & \textcolor{green}{0.687$\pm$0.064} \\
SSIM $\uparrow$ & 0.836$\pm$0.113 & \textcolor{green}{0.986$\pm$0.037} & \textcolor{red}{1.019$\pm$0.048} & 0.999$\pm$0.039 & 0.918$\pm$0.044 & 0.986$\pm$0.035 \\
\bottomrule
\end{tabular}
\end{table*}

\begin{table*}[htbp]
\caption{Performance Comparison of Fusion Methods on the LLVIP Dataset}
\label{tab:llvip_results}
\centering
\begin{tabular}{lccccc}
\toprule
\textbf{Method} & \textbf{DetFusion} & \textbf{SeAFusion} & \textbf{CDDFuse} & \textbf{PSFusion} & \textbf{PIAFusion} \\
\midrule
Speed $\downarrow$ & \textcolor{green}{0.133} & \textcolor{red}{0.100} & 1.640 & 0.283 & 0.189 \\
EN $\uparrow$ & 7.307$\pm$0.324 & \textcolor{green}{7.405$\pm$0.195} & 7.332$\pm$0.244 & \textcolor{red}{7.675$\pm$0.101} & 7.373$\pm$0.187 \\
SD $\uparrow$ & \textcolor{red}{67.523$\pm$6.721} & 49.576$\pm$7.957 & 49.095$\pm$10.199 & \textcolor{green}{51.162$\pm$13.675} & 50.197$\pm$8.505 \\
MI $\uparrow$ & 2.677$\pm$0.256 & 4.067$\pm$0.481 & \textcolor{red}{4.516$\pm$0.647} & 3.168$\pm$0.364 & \textcolor{green}{4.203$\pm$0.577} \\
PSNR $\uparrow$ & 63.41$\pm$1.18 & 62.257$\pm$1.49 & \textcolor{red}{62.895$\pm$1.019} & 60.373$\pm$1.007 & \textcolor{green}{62.274$\pm$1.252} \\
Qabf $\uparrow$ & 0.573$\pm$0.039 & \textcolor{red}{0.618$\pm$0.067} & \textcolor{green}{0.622$\pm$0.072} & 0.595$\pm$0.046 & 0.674$\pm$0.05 \\
SSIM $\uparrow$ & 0.738$\pm$0.052 & \textcolor{green}{0.938$\pm$0.043} & \textcolor{red}{0.935$\pm$0.036} & 0.800$\pm$0.053 & 0.924$\pm$0.047 \\
\bottomrule
\end{tabular}
\end{table*}
Table~\ref{tab:msrs_results} and Table~\ref{tab:llvip_results} report the quantitative performance of the fusion algorithms on the MSRS and LLVIP datasets.

On the MSRS dataset, which consists of diverse urban scenes in both daytime and nighttime conditions, PSFusion achieves the highest EN (6.838) and SD (46.279), indicating strong contrast and richness in details. DetFusion records the highest PSNR (65.26), reflecting high reconstruction fidelity, while CDDFuse leads in mutual information (4.951), capturing more complementary information. SeAFusion shows the best SSIM (0.986), and PIAFusion performs competitively across multiple metrics with strong MI and Qabf scores, while maintaining a fast inference speed (0.223 s).

On the LLVIP dataset, which focuses on extremely low-light conditions with pedestrian-centric annotations, PSFusion again achieves the highest EN (7.675), effectively enhancing salient targets. However, DetFusion reaches the highest SD (67.523), though at the cost of some structural quality (SSIM: 0.738). CDDFuse delivers the best PSNR (62.895) and MI (4.516), while SeAFusion and PIAFusion both perform strongly in SSIM (0.938 and 0.924, respectively), demonstrating good structural preservation. SeAFusion also achieves the fastest fusion speed (0.100 s), followed by DetFusion and PIAFusion.

\subsection{Results on Downstream Tasks}
To further investigate the performance of fused images in high-level computer vision tasks, we conduct pedestrian detection experiments using both the fused images and the original infrared and visible source images. The experiments are performed on the campus dual-spectrum dataset and the LLVIP dataset, focusing on human detection under varying illumination conditions.

We adopt the Lang-Segment-Anything framework for object detection, which enables language-driven segmentation and detection. In our experiments, the \texttt{text-threshold} and \texttt{iou-threshold} parameters are set to 0.3 and 0.5, respectively.

\subsubsection{Qualitative Results}
The campus dual-spectrum dataset includes two representative scenarios: daytime and nighttime. To comprehensively evaluate the detection performance, we conduct pedestrian detection experiments under three settings: the full dataset, daytime only, and nighttime only.

\begin{figure}[htbp]
    \centering
    \includegraphics[width=1\linewidth]{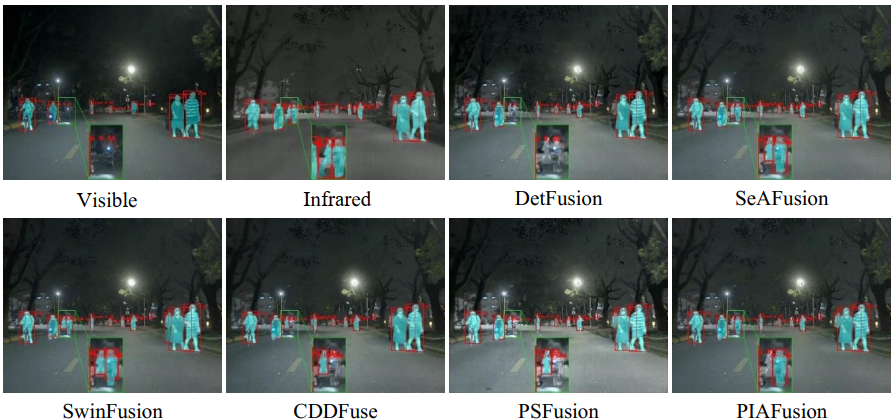}
    \caption{Results on '000053' of campus dataset.}
    \label{fig:campus_night_1}
\end{figure}

\begin{figure}[t]
    \centering
    \includegraphics[width=1\linewidth]{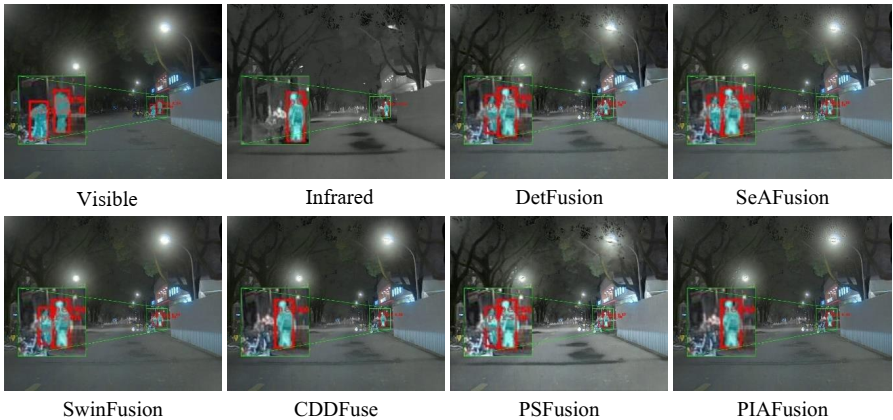}
    \caption{Results on '000044' of campus dataset.}
    \label{fig:campus_night_2}
\end{figure}

Fig.~\ref{fig:campus_night_1} and Fig.~\ref{fig:campus_night_2} qualitatively illustrate the detection results of different baseline methods on two representative nighttime scenes from the campus dataset. Compared with single-modality inputs, fused images demonstrate two notable advantages.
First, they effectively preserve distinguishable targets captured by the infrared modality, enabling more reliable detection under poor lighting conditions.

\begin{figure}
    \centering
    \includegraphics[width=1\linewidth]{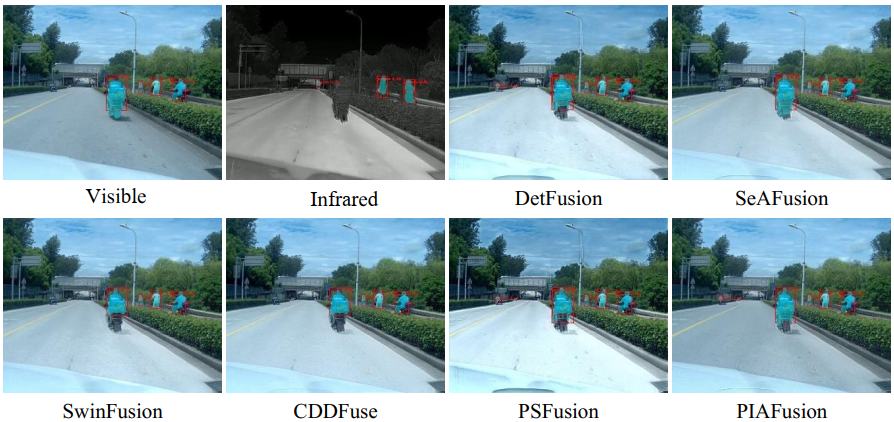}
    \caption{Results on '000105' of campus dataset.}
    \label{fig:ride_a_motorbike}
\end{figure}

As shown in Fig.~\ref{fig:ride_a_motorbike}, pedestrians riding electric bicycles and those at a distance exhibit high contrast and clearly defined contours in the fused images, which greatly enhances visual interpretability. Compared with infrared-only images, fused images enable more accurate bounding box annotations, especially in challenging scenarios such as crowded scenes or distant pedestrians.
Moreover, in contrast to visible images, fused images can successfully detect pedestrians under low-light conditions while retaining the rich texture details from the visible modality—such as the striped short-sleeve shirt worn by a pedestrian—which better aligns with human visual perception.
Among the baseline methods, SeAFusion and SwinFusion particularly excel at highlighting distinguishable targets and preserving sharper edge contours. 

Compared with single-modality images, fused images enable the detection of all relevant targets in the scene, thereby improving overall detection accuracy. For example, methods such as DetFusion, PSFusion, and PIAFusion successfully detect the person driving an electric tricycle on the left side of the image—a target that is missed or incorrectly identified in both infrared and visible images individually.
This highlights the superiority of fusion-based approaches in reducing false positives and recovering missed detections, particularly in complex or low-visibility scenes.

\subsubsection{Quantitative Analysis}
We adopt mAP@50 (mean Average Precision at IoU threshold 0.5) as the quantitative evaluation metric for object detection performance. Table~\ref{tab:map_results} presents the detection results of the fused images and single-modality (infrared and visible) images on the campus dual-spectrum dataset.
\begin{table}[htbp]
\caption{Detection Performance (mAP@50) on the Campus Dual-Spectrum Dataset}
\label{tab:map_results}
\centering
\begin{tabular}{lccc}
\toprule
\textbf{Method} & \textbf{All} & \textbf{Day} & \textbf{Night} \\
\midrule
IR & 0.5735 & 0.4456 & 0.6705 \\
RGB & 0.6043 & 0.6741 & 0.5180 \\
CDDFuse & 0.7150 & \textcolor{red}{0.7352} & 0.6838 \\
SeAFusion & 0.7153 & \textcolor{green}{0.7338} & \textcolor{green}{0.7084} \\
PIAFusion & 0.7081 & 0.7247 & 0.6826 \\
PSFusion & \textcolor{green}{0.7359} & 0.7335 & \textcolor{red}{0.7125} \\
DetFusion & \textcolor{red}{0.7364} & 0.7312 & \textcolor{red}{0.7125} \\
SwinFusion & 0.7050 & 0.6681 & 0.7048 \\
\bottomrule
\end{tabular}
\end{table}
It can be observed that nearly all fusion-based methods achieve significantly better detection accuracy compared to their single-modality counterparts. This confirms that image fusion can effectively leverage the complementary information from both modalities to enhance detection performance. Notably, CDDFuse and SeAFusion achieve superior detection accuracy in daytime scenes, while PSFusion, DetFusion, and SeAFusion perform better under nighttime conditions. Overall, PSFusion and DetFusion exhibit the best detection performance across all conditions in the campus dual-spectrum dataset.

This is consistent with the findings from the qualitative analysis: the fused images generated by PSFusion and DetFusion maintain high contrast, effectively incorporate thermal radiation information from infrared images, highlight hard-to-observe targets, and preserve the rich texture details from the visible modality.
\begin{table}[htbp]
\caption{Detection Performance (mAP@50) on the LLVIP Dataset}
\label{tab:llvip_map}
\centering
\resizebox{\linewidth}{!}{
\begin{tabular}{lcccccc}
\toprule
\textbf{Method} & \textbf{IR} & \textbf{RGB} & \textbf{CDDFuse} & \textbf{SeAFusion} & \textbf{PIAFusion} & \textbf{PSFusion} \\
\midrule
mAP50 & 0.8546 & 0.7929 & \textcolor{red}{0.8563} & \textcolor{green}{0.8560} & 0.8394 & 0.8527 \\
\bottomrule
\end{tabular}}
\end{table}
To further investigate the effectiveness of fusion methods, Table~\ref{tab:llvip_map} presents the quantitative detection results (mAP@50) of both fused and single-modality images on the LLVIP dataset. Among all methods, CDDFuse and SeAFusion achieve the best detection performance. CDDFuse effectively integrates edge and contour information from the source modalities, enhancing the model’s perception of object boundaries and contributing to more precise segmentation and detection. This demonstrates the potential of structure-aware fusion strategies in improving high-level vision tasks under low-light conditions.

In summary, image fusion significantly improves object detection performance. It is worth noting that fused images which preserve source texture details, edge and contour information, and maintain high contrast for salient targets tend to yield better results in detection tasks.

\section{Findings}
Based on extensive experiments and analyses conducted on three benchmark datasets (MSRS, LLVIP, and the proposed all-weather campus dual-spectrum dataset), we derive the following key findings:

1. Fusion Significantly Enhances Detection Performance
Compared with single-modality inputs, fused images show clear advantages in object detection tasks. They effectively combine the thermal radiation information of infrared images with the fine-grained textures of visible images, which improves the visibility and localization of targets under both normal and challenging conditions (e.g., low light, occlusion, and glare).

2. General Metrics Are Not Sufficient for Task Performance Assessment
Algorithms that perform well under standard fusion metrics (e.g., EN, SD, SSIM, PSNR) do not necessarily yield superior results in downstream tasks such as object detection. This discrepancy highlights the limitations of relying solely on general-purpose metrics and underscores the importance of incorporating task-driven evaluation to better reflect practical application performance.

3. Task-Oriented Fusion Design Yields Consistent Gains
Algorithms explicitly designed with downstream tasks in mind—such as DetFusion (for detection) and SeAFusion (for segmentation)—demonstrate robust and balanced performance across both general fusion metrics and detection results. This validates the effectiveness of integrating high-level vision supervision into the fusion process.

4. High Contrast and Edge Preservation Are Critical
Fusion methods that emphasize contrast enhancement, edge clarity, and structural preservation—such as PSFusion and CDDFuse—tend to perform better in both qualitative and quantitative evaluations. These attributes are particularly beneficial in improving model perception of object boundaries, which contributes to more accurate detection and recognition.

5. Dataset Conditions Affect Method Suitability
While CDDFuse and SeAFusion perform best in daytime or well-lit scenes, PSFusion and DetFusion achieve stronger results under nighttime or low-light conditions. This suggests that adaptive fusion strategies or hybrid frameworks may further improve performance across diverse real-world scenarios.

\section{Conclusions}
This paper presents a comprehensive evaluation of six recent visible-infrared image fusion algorithms across both general metrics and downstream object detection tasks. To support more realistic evaluation, we also introduce an all-weather campus dual-spectrum dataset covering diverse lighting conditions.

Experimental results demonstrate that fused images significantly outperform single-modality inputs in detection accuracy, especially under low-light and occluded scenarios. Algorithms incorporating high-level task supervision, such as DetFusion and SeAFusion, consistently achieve strong performance across both fusion quality and detection effectiveness. These findings underscore the limitations of traditional fusion metrics and advocate for task-driven evaluation and design.

In summary, integrating downstream task objectives into the fusion process is a promising direction to improve the real-world utility of multi-modal vision systems.

\vfill

\clearpage

\bibliographystyle{IEEEtran}
\bibliography{reference}

\end{document}